\documentclass{article}
\usepackage{spconf,amsmath,amssymb,graphicx}

\usepackage{booktabs}
\usepackage{multirow}
\usepackage{makecell}
\usepackage{placeins}

\usepackage[hidelinks]{hyperref}

\newcommand{\tabfont}{\fontsize{9}{9.2}\selectfont}
\newcommand{\bb}{\mathbf{b}}
\newcommand{\pp}{\mathbf{p}}
\newcommand{\tvec}{\mathbf{t}}
\newcommand{\sg}{\operatorname{sg}}
\newcommand{\ap}[1]{\text{AP}_{#1}}

\title{Oriented Semi-DETR for Semi-Supervised Aerial Object Detection}

\name{Yuqiu Li$^1$ \qquad Jian Xue$^1$ \qquad Fan Li$^1$ \qquad Ke Lu$^{1,2,\dagger}$
\thanks{$^\dagger$\emph{Corresponding author: Ke Lu}}}
\address{$^{1}$University of Chinese Academy of Sciences \quad $^{2}$Peng Cheng Laboratory}

\begin{document}
\ninept
\maketitle

\begin{abstract}
Semi-supervised object detection has recently been extended to DETR, where
oriented boxes introduce an ambiguity that does not arise for axis-aligned
boxes: the map from parameters to rectangles is two-to-one, since
$(c_x,c_y,w,h,\theta)$ and
$(c_x,c_y,h,w,\theta+\pi/2)$ denote the same rotated rectangle. Losses and
matching costs are computed from the parameters, so two boxes that coincide in
space can be far apart in the metric that drives training. A correctly localized
student box is then penalized by the regression loss, and the inflated matching
cost leads the Gaussian mixture model (GMM) filter to discard it. We propose
Oriented Semi-DETR, which removes this ambiguity wherever a pseudo box enters
training. Swap-aligned Target Selection regresses towards the encoding closer to
the student prediction. A Swap-invariant Reliability Cost minimizes over the two
encodings inside the matching cost, so the filter decides on geometry alone.
Swap-aware Pseudo-label Denoising starts denoising earlier, ramps its loss
weight linearly, and weights each query by the teacher score and encoding margin.
All three are parameter-free and require no additional forward pass. Across all label ratios
of DOTA-v2.0, Oriented Semi-DETR outperforms DETR-based supervised and
semi-supervised baselines, with the largest gains at the strict IoU threshold.
\end{abstract}


\section{Introduction}

Aerial images contain objects at arbitrary orientations, so detectors in this
domain predict oriented bounding boxes (OBBs) rather than axis-aligned boxes
\cite{ding2019roitransformer,xie2021orientedrcnn}. Annotation is expensive: a
single DOTA image may contain hundreds of instances, and each one needs four
vertices instead of two corners \cite{xia2018dota,ding2022dotav2}.
Semi-supervised object detection is therefore attractive in this domain,
provided that pseudo-label noise does not become entangled with the box
representation.

Teacher--student frameworks generate pseudo labels from a teacher view and train
a student on a stronger augmentation \cite{sohn2020stac,xu2021softteacher}, and
an exponential moving average keeps the teacher stable
\cite{tarvainen2017mean,liu2021unbiased,xu2021softteacher}. Detection
transformers assign one query to each object
\cite{carion2020detr,zhu2021deformable,zhang2023dino}. That assignment becomes
unstable once the targets are pseudo labels, so such detectors are hard to train
under semi-supervision \cite{zhang2023semidetr}. Semi-DETR
\cite{zhang2023semidetr} addresses this with stage-wise hybrid matching,
cross-view query consistency, and a Gaussian mixture model (GMM) that separates
reliable from unreliable pseudo boxes by their Hungarian cost. Sparse Semi-DETR
\cite{shehzadi2024sparse} adds sparse learnable queries. Neither was designed
for oriented boxes, and neither resolves their two equivalent encodings.

The obstacle is that the map from OBB parameters to rectangles is two-to-one.
Let $\sigma$ exchange $w$ and $h$ and rotate $\theta$ by $\pi/2$; then $\bb$ and
$\sigma(\bb)$ cover the same pixels and are the two encodings of that
rectangle. Supervised oriented detectors
handle the resulting loss discontinuity with angular encodings or
distribution-based losses
\cite{yang2020csl,yang2021gwd,yang2021kld,yu2023psc}, and aerial-image
transformers follow the same practice \cite{zeng2024arsdetr}. These methods
assume a fixed annotation convention. Under semi-supervision the target is a
teacher output, so the teacher and the student may adopt different encodings. A student query may reproduce the pseudo box geometrically and still
receive a large penalty that the pipeline cannot distinguish from a genuine
localization error. Recent semi-supervised oriented detectors weight the
consistency loss by the teacher--student orientation difference
\cite{liang2025soodpp} or combine several agreement cues \cite{wang2025multi}.
Neither is built on a detection transformer, so neither has a Hungarian
assignment or denoising queries, the two places our method operates on.
The ambiguity matters most when pseudo labels first become useful: at 20k
iterations the swapped encoding is closer for 16.9\% of matched pseudo-label and
query pairs. Replacing
the fixed regression cost with a swap-invariant one changes 18.9\% of Hungarian
assignments and 7.0\% of GMM filtering decisions. The effect concentrates on near-square boxes and weakens
as the teacher matures, by which time it has already shaped the
pseudo-label distribution.

\begin{figure*}[t]
  \centering
  \includegraphics[width=0.98\textwidth]{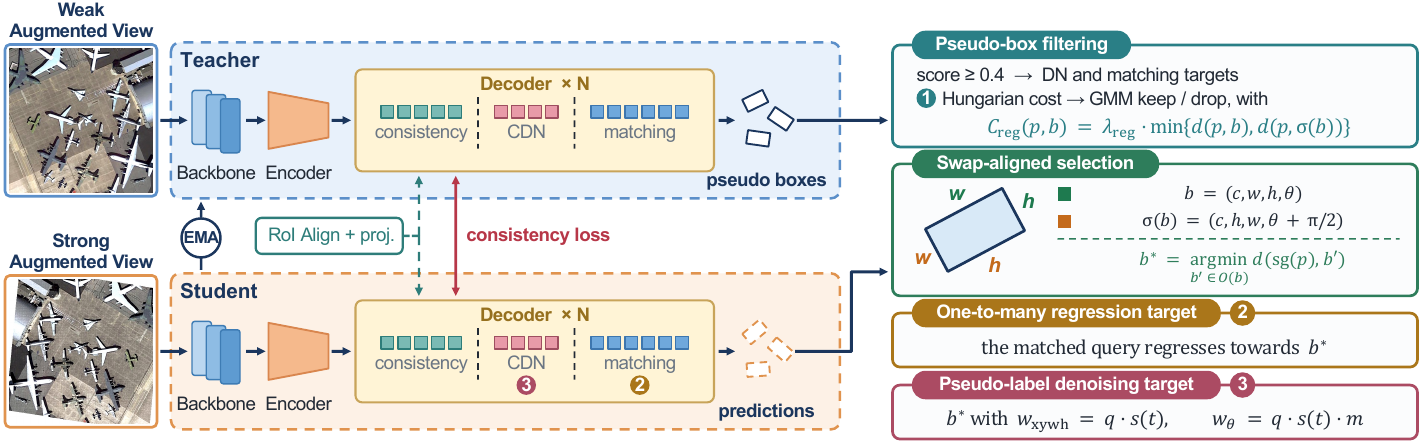}
  \vspace{-10pt}
  \caption{Overview of Oriented Semi-DETR. The teacher runs on the weakly
  augmented view and produces pseudo boxes, and the student runs on the
  strongly augmented view. Both decoders receive the same three query groups,
  kept apart by an attention mask: consistency queries, contrastive denoising
  (CDN) queries, and matching queries. An oriented box has two encodings,
  $\bb$ and $\sigma(\bb)$, for the same rectangle. We select the one closer to
  the student prediction and apply that choice at the numbered steps.}
  \label{fig:overview}
  \vspace{-17pt}
\end{figure*}

Oriented Semi-DETR comprises three components, summarized in
Fig.~\ref{fig:overview}. First, Swap-aligned Target Selection regresses towards
whichever encoding is closer to the student prediction, at the one-to-many
regression target and at the supervised and pseudo-label denoising targets. The
selection is a hard $\arg\min$, so the target is always a valid rectangle.
Second, a Swap-invariant Reliability Cost takes the minimum over the two
encodings within the matching cost, so the filter decides on geometry
rather than on the encoding the teacher produced. Third, Swap-aware Pseudo-label
Denoising advances the start of pseudo-label denoising, ramps its loss weight
linearly, and scales each query by the teacher score $q$ and the encoding margin
$m$.

\vspace{-10pt}
\section{Method}

\subsection{Preliminaries}

We build on Semi-DETR \cite{zhang2023semidetr} with a DINO
\cite{zhang2023dino} detector adapted to oriented boxes. A teacher and a student
share the architecture, the teacher is the exponential moving average of the
student, and each iteration samples one labeled and one unlabeled image. The
teacher runs on the weakly augmented view and its outputs above a score
threshold become pseudo boxes. Every pseudo box is matched to a student query by
a Hungarian assignment whose cost combines classification, regression and
rotated IoU terms. A two-component GMM is fitted to the resulting costs and
splits the pseudo boxes into a reliable and an unreliable set. Denoising (DN)
queries follow DN-DETR's perturbation of ground-truth boxes
\cite{li2022dndetr}; in the unlabeled branch they are built from teacher pseudo
boxes, following Semi-DETR \cite{zhang2023semidetr}. Each decoder therefore
receives three query groups, separated by an attention mask: consistency, contrastive denoising, and
matching queries, as shown in Fig.~\ref{fig:overview}.
An oriented box is written $\bb=(c_x,c_y,w,h,\theta)$. We store the angle
normalized as $a=\theta/\pi+1/2\in[0,1)$ and the center and size normalized by
the image width and height.

\vspace{-5pt}
\subsection{The two encodings of an oriented box}

Define
\begin{equation}
  \sigma(c_x,c_y,w,h,\theta)=\bigl(c_x,c_y,h,w,\operatorname{wrap}(\theta+\pi/2)\bigr),
  \label{eq:sigma}
\end{equation}
where $\operatorname{wrap}$ returns the angle to $[-\pi/2,\pi/2)$. Applying
$\sigma$ twice returns the original box, so every rectangle corresponds to the
pair $\{\bb,\sigma(\bb)\}$, whose members differ only in which side is called
the width. One implementation detail matters: $w$ and $h$ are normalized by different
image dimensions, so the exchange must be performed on absolute pixel values and
the result normalized again. Exchanging the normalized values directly gives a
different rectangle unless the image is square.

We measure the disagreement between a prediction $\pp$ and a target $\tvec$ with
\begin{equation}
  d(\pp,\tvec)=|\Delta c_x|+|\Delta c_y|+|\Delta w|+|\Delta h|
             +\lambda\,\varphi(\Delta a),
  \label{eq:dist}
\end{equation}
where every term is taken on normalized coordinates,
$\varphi(\Delta a)=\min(|\Delta a|,1-|\Delta a|)\in[0,1/2]$ is the wrapped
distance on the unit circle, and $\lambda=1$. The same $\varphi$ and $\lambda$
are shared by the regression loss, the matching cost and the selection rule of
Section~\ref{sec:select}, so within one configuration boxes are never compared
under two different metrics.

\vspace{-5pt}
\subsection{Swap-aligned target selection}
\label{sec:select}

Let $\pp$ be a student prediction assigned to a target $\tvec$. Instead of
regressing towards whichever encoding the teacher produced, we regress towards
\begin{equation}
  \tvec^{*}=\arg\min_{\tvec'\in\{\tvec,\sigma(\tvec)\}}\; d\bigl(\sg(\pp),\tvec'\bigr),
  \label{eq:select}
\end{equation}
where $\sg$ denotes the stop-gradient and a tie is resolved in favor of the
teacher's own encoding. The selection never interpolates between the two
encodings, since an interpolated box would not be a valid rectangle. Writing
$d_0=d(\pp,\tvec)$ and $d_1=d(\pp,\sigma(\tvec))$ gives
$d(\pp,\tvec^{*})=\min(d_0,d_1)\le d_0$, so the selected target is never farther
from the prediction than the teacher's encoding, and the two coincide whenever
the teacher's encoding is already the closer one.

Equation~(\ref{eq:select}) is applied to three regression targets: that of the
queries selected by the one-to-many assigner, which carries most of the
gradient, and those of the supervised and pseudo-label DN queries. On the two
DN paths a perturbed copy of a box enters the decoder, which must recover
the original, and the ambiguity is most harmful there, since the perturbation is
small by construction and an encoding mismatch then dominates it. Keeping the
two DN paths separately switchable allows us to attribute them in
Section~\ref{sec:ablation}.

\vspace{-5pt}
\subsection{Swap-invariant reliability cost}

The GMM filter of Semi-DETR \cite{zhang2023semidetr} decides which pseudo boxes
survive, so the same ambiguity has a different consequence there: a pseudo
box may be discarded because its cost is inflated by the encoding. The matching
cost is a weighted sum of a classification, a regression and a rotated-IoU term,
and the GMM is fitted to the total cost of the matched pairs. We replace its
regression term by
\begin{equation}
  C_\text{reg}(\pp,\bb)=\lambda_\text{reg}\min\bigl(d(\pp,\bb),\,d(\pp,\sigma(\bb))\bigr).
  \label{eq:cost}
\end{equation}
Here $\lambda_\text{reg}$ is the regression weight of the matcher, set to $5$.
The penalty is substantial. The fixed matcher we compare against uses a cosine
angle penalty. If the student predicts exactly $\sigma(\bb)$, that term alone
contributes $\lambda_\text{reg}(1-\cos\pi)=10$, and the exchanged $w$ and $h$
add a further $2\lambda_\text{reg}|w-h|$. A box whose normalized width and
height differ by $0.15$ therefore costs about $11.5$ although the student has
localized it perfectly, while the rotated IoU term of the same matching cost
reports a perfect overlap. Equation~(\ref{eq:cost}) returns $0$ for the same
pair. Because
the GMM is refitted at every iteration, inflating one part of the cost
distribution moves the decision boundary for every pseudo box, including
unambiguous ones.

\vspace{-6pt}
\subsection{Swap-aware pseudo-label denoising}

Denoising perturbs a target box and requires the decoder to recover it. On the
labeled branch the target is an annotation, so every denoising query carries the
same amount of information. On the unlabeled branch the target is a pseudo box
whose reliability changes from one instance to the next. We therefore separate
the activation schedule from the instance-wise weighting. Let $q_j\in[0,1]$ be the
teacher score of the pseudo box behind the $j$-th denoising query. We use a
linear ramp
\begin{equation}
  s(t)=\min\Bigl(\max\Bigl(\frac{t-t_0}{T},0\Bigr),1\Bigr),
  \label{eq:conf}
\end{equation}
where $t$ is the current iteration, $t_0$ is the activation step and $T$ is the
ramp length. The pipeline we build on enables pseudo-label denoising at
160k, at the second decay of the learning rate. We instead set $t_0=120$k and
$T=10$k, so denoising begins at the first decay and its weight rises over 10k
iterations. We refer to this schedule as the early ramp. The first setting applies the ramp alone, so
every denoising query of the iteration carries $w_j=s(t)$. The second scales the queries by the teacher score,
\begin{equation}
  w_{\mathrm{xywh},j}=q_j s(t),\qquad
  w_{\theta,j}^{\mathrm{score}}=q_j s(t).
  \label{eq:qdn}
\end{equation}

Oriented Semi-DETR adds an encoding margin to the angle regression. For the current student DN
prediction $\pp_j$ and pseudo box $\bb_j$, we define
\begin{equation}
  m_j=\frac{\left|d(\pp_j,\bb_j)-d(\pp_j,\sigma(\bb_j))\right|}
  {d(\pp_j,\bb_j)+d(\pp_j,\sigma(\bb_j))+\varepsilon},
  \qquad
  w_{\theta,j}^{\mathrm{margin}}=q_j s(t)m_j .
  \label{eq:margin}
\end{equation}
Here $\varepsilon=10^{-6}$, and $m_j\in[0,1)$ measures the relative separation
of the two encodings under the same distance $d$. It is small when the
prediction is nearly indifferent between them and large when one is clearly
closer. The factor $q_js(t)$ weights the
four spatial L1 coordinates and the per-box rotated-IoU term, while $m_j$
multiplies only the angle L1 coordinate, so it changes neither the
classification nor the IoU weight. All weights are computed without gradient,
and the target of Equation~(\ref{eq:select}) is unchanged.

\vspace{-5pt}
\subsection{Relation to prior treatments of the ambiguity}

A Gaussian-based loss maps a box to a two-dimensional Gaussian, and $\bb$ and
$\sigma(\bb)$ give the same Gaussian, so GWD \cite{yang2021gwd} and KLD
\cite{yang2021kld} are already invariant under $\sigma$. Substituting such a
loss changes the metric everywhere, including the labeled branch, where the
annotation convention fixes the encoding and the ambiguity never arises.
Equation~(\ref{eq:select}) leaves the metric unchanged and modifies the target
alone, so both branches remain under a single loss and supervised behavior is
preserved. The same $\arg\min$ extends to the GMM filter, which is a decision
rule rather than a loss.

RIDet \cite{ming2021optimization} is the closest supervised counterpart. Its
representation invariance loss also minimizes over the equivalent encodings of a
box, so its target is selected during training. Two differences
separate it from Equation~(\ref{eq:select}). Its target is an annotation whose
encoding a labeling convention already fixes, while ours is a teacher
output that obeys no convention and shifts as the teacher is updated. Its
minimization is also confined to the loss, so it does not reach the Hungarian
cost, the reliability filter, or the denoising queries built from pseudo boxes,
the three places where a semi-supervised pipeline encounters the ambiguity.

\begin{table*}[t]
  \centering
  \tabfont
  \caption{The results on DOTA-v2.0 test-dev. Oriented Semi-DETR uses the full configuration of
  Equations~(\ref{eq:conf})--(\ref{eq:margin}). The last group uses the whole trainval set as labeled data.}
  \label{tab:main}
  \vspace{1pt}
  \setlength{\tabcolsep}{2.6pt}
  \begin{tabular}{@{}lccc|ccc|ccc|ccc@{}}
    \toprule
    \multirow{2}{*}{Method}
      & \multicolumn{3}{c}{10\% labeled}
      & \multicolumn{3}{c}{20\% labeled}
      & \multicolumn{3}{c}{30\% labeled}
      & \multicolumn{3}{c}{Fully labeled} \\
    \cmidrule(lr){2-4}\cmidrule(lr){5-7}\cmidrule(lr){8-10}\cmidrule(lr){11-13}
      & $\ap{50}$ & $\ap{75}$ & mAP
      & $\ap{50}$ & $\ap{75}$ & mAP
      & $\ap{50}$ & $\ap{75}$ & mAP
      & $\ap{50}$ & $\ap{75}$ & mAP \\
    \midrule
    DINO \cite{zhang2023dino}
      & 40.72 & 21.76 & 22.49 & 46.31 & 24.94 & 25.82 & 48.11 & 26.71 & 27.16
      & 49.11 & 26.91 & 27.51 \\
    Semi-DETR \cite{zhang2023semidetr}
      & 43.94 & 25.62 & 25.82 & 48.01 & 28.10 & 28.10 & 50.63 & 29.47 & 29.60
      & 51.11 & 31.50 & 30.81 \\
    Sparse Semi-DETR \cite{shehzadi2024sparse}
      & 44.16 & 25.49 & 25.83 & 48.38 & 28.28 & 28.51 & 50.24 & 29.40 & 29.53
      & 51.71 & 32.26 & 31.22 \\
    \midrule
    Oriented Semi-DETR
      & \textbf{45.92} & \textbf{27.77} & \textbf{27.61}
      & \textbf{49.89} & \textbf{31.67} & \textbf{30.69}
      & \textbf{52.11} & \textbf{33.51} & \textbf{32.19}
      & \textbf{53.16} & \textbf{35.07} & \textbf{33.23} \\
    \bottomrule
  \end{tabular}
  \vspace{-15pt}
\end{table*}

\begin{table*}[t]
  \centering
  \tabfont
  \caption{The results on DOTA-v2.0 test-dev with pseudo-label denoising across label ratios. All rows use
  all three regression targets swap-aligned and successively add the early ramp,
  teacher-score weighting and encoding-margin weighting.}
  \label{tab:ratio-ablation}
  \vspace{1pt}
  \setlength{\tabcolsep}{2.6pt}
  \begin{tabular}{@{}lccc|ccc|ccc|ccc@{}}
    \toprule
    \multirow{2}{*}{Pseudo-label DN setting}
      & \multicolumn{3}{c}{10\% labeled}
      & \multicolumn{3}{c}{20\% labeled}
      & \multicolumn{3}{c}{30\% labeled}
      & \multicolumn{3}{c}{Fully labeled} \\
    \cmidrule(lr){2-4}\cmidrule(lr){5-7}\cmidrule(lr){8-10}\cmidrule(lr){11-13}
      & $\ap{50}$ & $\ap{75}$ & mAP
      & $\ap{50}$ & $\ap{75}$ & mAP
      & $\ap{50}$ & $\ap{75}$ & mAP
      & $\ap{50}$ & $\ap{75}$ & mAP \\
    \midrule
    \makecell[l]{Early pseudo-label DN\\ramp}
      & 45.56 & \textbf{27.88} & 27.52 & 49.19 & \textbf{32.15} & \textbf{30.69} & 51.84 & 33.57 & 31.96
      & 52.79 & 34.86 & 33.15 \\
    \makecell[l]{Early ramp +\\teacher score}
      & \textbf{45.92} & 27.77 & \textbf{27.65} & 49.64 & 31.95 & 30.50 & 52.08 & \textbf{33.60} & 32.18
      & 53.02 & 34.50 & 32.99 \\
    \makecell[l]{Early ramp + teacher score\\+ encoding margin}
      & \textbf{45.92} & 27.77 & 27.61 & \textbf{49.89} & 31.67 & \textbf{30.69} & \textbf{52.11} & 33.51 & \textbf{32.19}
      & \textbf{53.16} & \textbf{35.07} & \textbf{33.23} \\
    \bottomrule
  \end{tabular}
  \vspace{-20pt}
\end{table*}

\section{Experiments}

\subsection{Setup}

We merge the train and val sets of DOTA-v2.0
\cite{xia2018dota,ding2022dotav2} into a trainval set, follow the partial-label
protocol of Semi-DETR \cite{zhang2023semidetr} to sample the 10\%, 20\% and
30\% labeled splits from it, and add a fully labeled setting that uses the entire
trainval set as labeled data and the test-challenge images as unlabeled data,
after we exclude the frames under heavy cloud or fog and the frames whose
foreground is almost black. The fully labeled setting starts the unlabeled loss at 20k instead
of 10k and is otherwise identical. The detector is DINO \cite{zhang2023dino} with a ResNet-50
\cite{he2016resnet} backbone. Training runs for 180k iterations with AdamW at a
learning rate of $10^{-4}$, decayed at 120k and 160k. We set the EMA momentum to
$0.999$, the unsupervised loss weight to $0.5$, the pseudo-label score threshold
to $0.4$, and $\lambda_\text{reg}=5$. As in the
one-to-many-first design of Semi-DETR \cite{zhang2023semidetr}, the warm-up
length equals the full schedule, so every system in Table~\ref{tab:main} stays
in the one-to-many stage and the comparison is not confounded by the stage
switch. The target-selection ablation keeps the legacy 160k pseudo-label DN
start and the nested settings use $t_0=120$k and $T=10$k. All numbers are from
the official DOTA evaluation server on test-dev, and we report
$\ap{50}$, $\ap{75}$ and mean average precision (mAP). Under the partial-label
protocol, results for Oriented Semi-DETR are averaged over three training runs.

\vspace{-10pt}
\subsection{Main results}

Table~\ref{tab:main} compares Oriented Semi-DETR with the DETR-based supervised
and semi-supervised baselines at each label ratio. Against the best mAP among
them, the gains are 1.78, 2.18, and 2.59 points at 10\%, 20\%, and 30\%;
against the best $\ap{75}$ they are 2.15, 3.39, and 4.04 points. The larger gain
at the strict threshold is consistent with better alignment of matched boxes,
although aggregate metrics do not separate recall from localization.
Table~\ref{tab:ratio-ablation} decomposes the full configuration into the three
pseudo-label denoising components at every ratio, with all three regression
targets swap-aligned throughout. Teacher-score weighting gives the best mAP at
10\%. At 20\% the encoding margin restores $\ap{50}$ but reduces $\ap{75}$, leaving mAP
at the level of the ramp alone. At 30\% the full configuration leads on $\ap{50}$
and mAP. No component is best on every metric at every partial ratio, so we
retain the full configuration throughout.

With the entire trainval set used as labeled data, Oriented Semi-DETR reaches
53.16 $\ap{50}$, 35.07 $\ap{75}$ and 33.23 mAP. The same baseline is strongest on
all three metrics here, so the gains are 1.45, 2.81 and 2.01 points, widest once
more at the strict threshold. The step from the 30\% split is about one point of mAP,
and two factors limit it. The 180k schedule is the one used for the partial
splits and is short for a labeled set three times the size, and the
test-challenge images that serve as unlabeled data carry more noise and more
hard cases than the trainval set. The three denoising components also rank
consistently here, unlike at any partial ratio: the full configuration leads on
all three metrics, and the
encoding margin recovers the $\ap{75}$ that teacher-score weighting alone loses.

\vspace{-10pt}
\subsection{Ablation}
\label{sec:ablation}

Table~\ref{tab:ablation} isolates the three regression targets. The swap-invariant cost of
Equation~(\ref{eq:cost}) is active in every row, so the table measures the
target selection of Equation~(\ref{eq:select}) alone.

\begin{table}[t]
  \centering
  \tabfont
  \caption{Contribution of each regression target. The swap-invariant GMM cost is
  active in all rows, and the last column gives the pseudo-label denoising design.
  The results are from DOTA-v2.0 test-dev under 30\% labeled protocol.}
  \label{tab:ablation}
  \setlength{\tabcolsep}{3.8pt}
  \begin{tabular}{@{}ccclccc@{}}
    \toprule
    \makecell{Regr.\\target} & \makecell{Sup.\\DN} & \makecell{Unsup.\\DN}
      & \makecell{Pseudo-DN\\schedule}
      & $\ap{50}$ & $\ap{75}$ & mAP \\
    \midrule
             &            &            & --  & 49.15 & 28.68 & 28.69 \\
             & \checkmark &            & --  & 50.47 & 30.55 & 30.14 \\
    \checkmark & \checkmark &           & --  & 51.44 & 33.54 & 31.89 \\
    \checkmark & \checkmark & \checkmark & --  & 51.98 & 33.40 & 31.95 \\
    \midrule
    \checkmark & \checkmark & \checkmark & \makecell[l]{early ramp} & 51.84 & 33.57 & 31.96 \\
    \checkmark & \checkmark & \checkmark & \makecell[l]{early ramp $+\,q$} & 52.08 & \textbf{33.60} & 32.18 \\
    \checkmark & \checkmark & \checkmark & \makecell[l]{early ramp $+\,q+m$} & \textbf{52.11} & 33.51 & \textbf{32.19} \\
    \bottomrule
  \end{tabular}
  \vspace{-10pt}
\end{table}

Applying the rule to the supervised DN target alone yields $+1.9$ points of
$\ap{75}$. This supports the argument of Section~2.3 that denoising is the path
most exposed to the ambiguity, since the injected noise is small and an encoding
mismatch then dominates the correction the decoder must make. Adding the
ordinary regression target contributes a further 3.0 points, the largest single
step, as expected from the target that carries most of the gradient under
one-to-many assignment. The fourth row adds the unsupervised DN target, active
over the final 20k iterations. It gains 0.54 $\ap{50}$ and 0.06 mAP and loses
0.14 $\ap{75}$, so the gain appears at the loose threshold alone.

The last three rows add the changes of Section~2.5. Relative to the legacy 160k
schedule, the early ramp changes $\ap{75}$ by $+0.17$ and mAP by $+0.01$. Adding
the teacher score gives a further $+0.03$ $\ap{75}$ and $+0.22$ mAP. Adding the
encoding margin raises $\ap{50}$ by $0.03$ and mAP by $0.01$ but reduces $\ap{75}$
by $0.09$, so this screening does not establish an additional strict-IoU
benefit.

Table~\ref{tab:angle-ablation} varies the angle penalty $\varphi$ of
Equation~(\ref{eq:dist}) with everything else fixed at the configuration that
aligns all three regression targets. The wrapped form leads the cosine form by 2.15 $\ap{50}$, 1.81
$\ap{75}$ and 1.36 mAP, and raw L1 falls between the two. With $\lambda$ held
at $1$, the cosine row also differs in the effective weight of the angle term,
so the table identifies the better configuration rather than isolating the
shape. It concerns $\varphi$ alone and is therefore reported apart from
Table~\ref{tab:ablation}.

\begin{table}[t]
  \centering
  \tabfont
  \caption{Angle penalty $\varphi$ in Equation~(\ref{eq:dist}), with
  all three regression targets swap-aligned. The results are from 30\% labeled protocol.}
  \label{tab:angle-ablation}
  \begin{tabular}{@{}lccc@{}}
    \toprule
    $\varphi(\Delta a)$ & $\ap{50}$ & $\ap{75}$ & mAP \\
    \midrule
    $1-\cos(2\pi\Delta a)$ (cosine)      & 49.83 & 31.59 & 30.59 \\
    $|\Delta a|$ (raw L1)                & 51.53 & 33.04 & 31.76 \\
    $\min(|\Delta a|,1-|\Delta a|)$      & \textbf{51.98} & \textbf{33.40} & \textbf{31.95} \\
    \bottomrule
  \end{tabular}
  \vspace{-8pt}
\end{table}

\vspace{-5pt}
\subsection{Prevalence of the ambiguity}

Table~\ref{tab:diag} reports the diagnostic behind the claim in Section~1. We
take teacher pseudo boxes and raw student queries from a training checkpoint,
run the Hungarian assignment and the GMM filter twice, once with the fixed cost
and once with Equation~(\ref{eq:cost}), and compare the outcomes.

\begin{table}[t]
  \centering
  \tabfont
  \caption{Fixed vs. swap-invariant reliability cost on teacher
  outputs, 128 unlabeled images. ``Sigma'' is the fraction of matched pairs whose
  swapped encoding is closer, and the flip columns count pairs whose assignment
  or filter decision changes.}
  \label{tab:diag}
  \vspace{1pt}
  \begin{tabular}{ccccc}
    \toprule
    \makecell{Train\\step} & \makecell{Matched\\pairs} & Sigma
      & \makecell{Assign.\\flip} & \makecell{Filter\\flip} \\
    \midrule
    20k  & 4671 & 16.89\% & 18.88\% & 7.02\% \\
    180k & 3826 & 3.89\%  & 4.89\%  & 2.88\% \\
    \bottomrule
  \end{tabular}
  \vspace{-6pt}
\end{table}

The fixed matcher selects a swapped query in only 0.09\% of the pairs
at 20k, against 16.89\% for the swap-invariant matcher. Mispricing is not the only
consequence. The assignment is diverted away from the ambiguous pairs before the
filter is applied. Both rates fall as the teacher converges, so a measurement taken late in
training understates the effect, which is largest while the pseudo-label
distribution is being formed.

Splitting the same pairs by aspect ratio confines the effect to near-square
boxes. In the aspect-ratio range $(1,1.3]$ the swapped encoding is closer for 43.9\% of the
pairs, 45.5\% of the assignments change, and the retention rate rises from 28.6\% to
44.9\%, while above ratio 2 the first two numbers drop to 7.1\% and 8.6\%. The
overall retention rate therefore shifts by only 1.20 points at 20k, although
7.02\% of the individual decisions change: the swap-invariant cost retains a comparable number of
pseudo boxes and a different selection of them. Storage tanks, roundabouts and vehicles
seen from above make near-square objects common in aerial scenes
\cite{xia2018dota,ding2022dotav2}, so a mechanism that would be marginal
in general detection is important here.

\vspace{-3pt}
\FloatBarrier
\section{Conclusion}

The parameterization of an oriented box is two-to-one: $\bb$ and $\sigma(\bb)$
denote the same rectangle, so a pseudo label does not determine a unique
regression target. Because the training signal is computed from the parameters,
the ambiguity acts twice. The regression loss penalizes a student box that
coincides with the pseudo box in space, and the Gaussian mixture filter removes
that box from the unlabeled loss. Oriented Semi-DETR resolves the
ambiguity at both points. It aligns every regression target with the encoding the student
already predicts, minimizes over the two encodings inside the reliability
cost, and starts pseudo-label denoising earlier under teacher-score and
encoding-margin weights. The architecture and the per-iteration
computation are unchanged. On DOTA-v2.0 our method improves performance on every
label ratio over DETR-based supervised and semi-supervised baselines, and the
gain is largest at the strict IoU threshold. The same approach applies
wherever a teacher--student target representation admits several equivalent
encodings of one oriented object.

\clearpage
\section*{Acknowledgements}
This work was supported by the Science Fund for Creative Research Groups of the
National Natural Science Foundation of China (62521007) and the Joint Key
Project of the National Natural Science Foundation of China and China
Electronics Technology Group Corporation (U21B2049).

\bibliographystyle{IEEEbib}
\bibliography{refs}

\end{document}


\maketitle

This document collects the material that did not fit in the main paper: the
properties of the two encodings and their proofs, a worked example of the cost
inflation, an implementation pitfall in the swap itself, the complete set of
ablations we ran, the full diagnostic tables, and the training hyperparameters.
Section and equation numbers carry the prefixes A, B, \ldots and S1, S2,
\ldots\ Numbers without a prefix refer to the main paper.

\section{Notation and the angle metric}

We write an oriented box as $\bb=(c_x,c_y,w,h,\theta)$ with the angle taken in
$[-\pi/2,\pi/2)$. The center and the size are normalized by the image width and
height, and the angle is stored as $a=\theta/\pi+1/2\in[0,1)$. In these
coordinates the half-turn of Eq.~(1) becomes $a\mapsto a+1/2 \bmod 1$, so a swap
always moves the normalized angle by exactly $1/2$.

The distance of Eq.~(2) is
\begin{equation}
  d(\pp,\tvec)=|\Delta c_x|+|\Delta c_y|+|\Delta w|+|\Delta h|
             +\lambda\,\varphi(\Delta a),
  \label{eq:supdist}
\end{equation}
and we studied three choices of $\varphi$:
\begin{align}
  \varphi_{\cos}(\Delta a)  &= 1-\cos(2\pi\,\Delta a), \label{eq:phicos}\\
  \varphi_{\text{wrap}}(\Delta a) &= \min(|\Delta a|,\,1-|\Delta a|),
    \label{eq:phiwrap}\\
  \varphi_{\text{raw}}(\Delta a) &= |\Delta a|. \label{eq:phiraw}
\end{align}
Table~\ref{tab:phi} lists their values at the two angles that matter. The
first two are $\pi$-periodic in $\theta$, so they treat $a=0$ and $a=1$ as the
same angle. The third is not, and it assigns a full unit of penalty across the
wrap point. We include it because the angle difference is small most of the time once
the target is swap-aligned, which makes the defect less visible than one would
expect, but the defect is real.

\begin{table}[t]
  \centering
  \tabfont
  \caption{The three angle penalties. $\Delta a=1/4$ is a $45^\circ$
  disagreement and $\Delta a=1/2$ is the swap. $\lambda=1$ in every
  configuration.}
  \label{tab:phi}
  \begin{tabular}{@{}lcccc@{}}
    \toprule
    $\varphi$ & $\Delta a=1/4$ & $\Delta a=1/2$ & periodic & smooth \\
    \midrule
    cosine     & 1.00 & 2.00 & yes & yes \\
    wrapped L1 & 0.25 & 0.50 & yes & no  \\
    raw L1     & 0.25 & 0.50 & no  & no  \\
    \bottomrule
  \end{tabular}
\end{table}

Two remarks on Table~\ref{tab:phi}. First, $\lambda$ stays at $1$ for all three
forms, so the three variants differ in the scale of the angle term as well as
in its shape. At a $45^\circ$ disagreement the cosine penalty is four times the
wrapped one. Section~\ref{sec:phi} therefore compares three complete
configurations, and it leaves the shape of the function uncontrolled.
Second, the cosine form has a maximum at the swap, where its derivative
vanishes. That stationary point is unstable rather than attracting, so we do
not attribute the behavior of the cosine runs to it.

The same $\varphi$ and the same $\lambda$ are used by the regression loss, by
the Hungarian regression cost and by the selection rule, in every configuration
we report. No run compares boxes under two different metrics.

\section{Properties of the two encodings}

Write $R(\bb)\subset\mathbb{R}^2$ for the set of points covered by the
rectangle that $\bb$ encodes.

\subsection{$\sigma$ is an involution and the two encodings are distinct}

Applying Eq.~(1) twice exchanges $w$ and $h$ back and adds $\pi$ to the angle,
which the wrap removes, so $\sigma\circ\sigma=\mathrm{id}$ and $\sigma$ is an
involution. The two encodings are always distinct as
parameter tuples, including for squares: $\operatorname{wrap}(\theta+\pi/2)=\theta$
has no solution in $[-\pi/2,\pi/2)$. Hence $|\{\bb,\sigma(\bb)\}|=2$ for every $\bb$.

\subsection{$\sigma$ preserves the rectangle}

Let $u(\theta)=(\cos\theta,\sin\theta)$ and $v(\theta)=(-\sin\theta,\cos\theta)$
be the two edge directions, so that
\begin{equation}
  R(\bb)=\Bigl\{(c_x,c_y)+s\,u(\theta)+t\,v(\theta)\;:\;
    |s|\le\tfrac{w}{2},\,|t|\le\tfrac{h}{2}\Bigr\}.
  \label{eq:rect}
\end{equation}
A quarter turn gives $u(\theta+\pi/2)=v(\theta)$ and
$v(\theta+\pi/2)=-u(\theta)$. Substituting these into Eq.~(\ref{eq:rect}) for
$\sigma(\bb)$, whose half-extents are $h/2$ along $u(\theta+\pi/2)$ and $w/2$
along $v(\theta+\pi/2)$, reproduces the same point set, since the range of $s$
and $t$ is symmetric about zero. The wrap adds a multiple of $\pi$ when the
result leaves $[-\pi/2,\pi/2)$, and a half turn sends $u\mapsto-u$ and
$v\mapsto-v$, which again leaves Eq.~(\ref{eq:rect}) unchanged. Therefore
$R(\sigma(\bb))=R(\bb)$.

The map from parameters to rectangles is consequently two-to-one, and its
fibers are exactly the pairs $\{\bb,\sigma(\bb)\}$. An annotation protocol
selects one member per
fiber and can do so consistently. A teacher network has no such
constraint, which is the source of the problem the main paper addresses.

\subsection{The selection never increases the regression distance}

Let $d_0=d(\pp,\tvec)$ and $d_1=d(\pp,\sigma(\tvec))$. By definition
$\tvec^{*}$ of Eq.~(3) attains $\min(d_0,d_1)$, so
\begin{equation}
  d(\pp,\tvec^{*})=\min(d_0,d_1)\le d_0,
  \label{eq:nonincrease}
\end{equation}
with equality whenever the teacher's encoding is already the closer member. The
selection is therefore never worse than the ordinary target on the quantity the
loss measures, and it is inactive on the majority of pairs. Table~\ref{tab:diagfull}
gives the fraction on which it is active.

\subsection{The training signal no longer depends on the encoding}

Since $\{\sigma(\tvec),\tvec\}=\{\tvec,\sigma(\tvec)\}$, replacing the teacher output
$\tvec$ by $\sigma(\tvec)$ leaves the feasible set of Eq.~(3) unchanged and
therefore leaves $\tvec^{*}$ unchanged. The loss the student receives is
invariant under the teacher's choice of encoding. This is the property we are
after, and no reweighting of the two candidates achieves it, because a weighted
target still moves when the weights are applied to a different member.

\subsection{Why a hard $\arg\min$}

Averaging the two encodings is not viable, because the pair is not convex in
parameter space. Take $\bb=(0,0,4,1,0)$, so that
$\sigma(\bb)=(0,0,1,4,\pi/2)$. The midpoint of the two is
$(0,0,2.5,2.5,\pi/4)$, a square of area $6.25$ turned by $45^\circ$. It shares
neither the shape nor the area of the original rectangle, whose area is $4$. Any
soft combination of the two encodings gives a box that is not the annotated
object.

\subsection{Cost of the rule}

$\tvec^{*}$ is computed from $\sg(\pp)$, so $\partial\tvec^{*}/\partial\pp=0$
and the backward pass is that of the ordinary loss with a constant target. The
rule introduces no parameter, no additional forward pass and no change to the
architecture. Its runtime cost is one extra evaluation of Eq.~(\ref{eq:supdist})
and one comparison per target. Inference is unchanged.

\section{A worked example of the cost inflation}

Section~2.4 states that the fixed regression cost reaches about $11.5$ for a
perfectly localized box. The arithmetic is as follows. Take a pseudo box whose
normalized width and height are $w=0.25$ and $h=0.10$, which is a
$256\times102$ pixel object in a $1024\times1024$ crop, and a student query that
predicts exactly $\sigma(\bb)$. The two boxes are the same rectangle, so the
student has localized the object without error. The individual terms of the
fixed cost are listed in Table~\ref{tab:cost}.

\begin{table}[t]
  \centering
  \tabfont
  \caption{Term-by-term breakdown of the fixed Hungarian regression cost for a
  student query that predicts the swapped encoding of a perfectly localized
  pseudo box. The cost weight is $5$ and $\lambda=1$, matching our
  configurations. Eq.~(4) evaluates the same pair at $0$, because the minimum
  over the two encodings is attained at $\sigma(\bb)$, which the student predicts
  exactly.}
  \label{tab:cost}
  \begin{tabular}{@{}lc@{}}
    \toprule
    Term & Fixed cost \\
    \midrule
    $|\Delta c_x|+|\Delta c_y|$ & $0$ \\
    $|\Delta w|+|\Delta h|$     & $0.30$ \\
    $\lambda\,\varphi_{\cos}(\Delta a)$, $\Delta a=1/2$ & $2.00$ \\
    \midrule
    $d$                         & $2.30$ \\
    $C_\text{reg}=5\,d$         & $\mathbf{11.5}$ \\
    \bottomrule
  \end{tabular}
\end{table}

The angle term alone contributes $5(1-\cos\pi)=10$, and the exchanged size adds
$5\times0.30=1.5$. The rotated IoU term of the same matching cost treats the two
encodings as one rectangle and reports a perfect overlap, so the fixed cost is
internally inconsistent: one of its terms reports an exact match and another
reports the worst possible one. Eq.~(4) returns $0$ and removes the
contradiction.

The consequence for the filter is not confined to the ambiguous pairs. Semi-DETR
\cite{zhang2023semidetr} fits a two-component GMM to the matched costs of each iteration and keeps the
pseudo boxes assigned to the low-cost component. A subset of costs inflated by
roughly $11.5$ forms a heavy right tail, which pulls the fitted components apart
and moves the decision boundary for every pseudo box in the batch. This is why
Table~\ref{tab:diagfull} shows a large number of filter flips while the total
number of retained boxes changes only slightly.

\section{The swap under anisotropic normalization}

The exchange of $w$ and $h$ must be carried out on absolute pixel values.
Doing it on the normalized values is wrong whenever the image is not square,
and the failure mode is subtle enough to deserve a note.

Let the image be $W\times H$ and write $\bar w=w/W$ and $\bar h=h/H$. Swapping
the normalized values gives back the pixel sizes
\begin{equation}
  w'=\bar h\,W=h\,\frac{W}{H},\qquad h'=\bar w\,H=w\,\frac{H}{W}.
  \label{eq:naiveswap}
\end{equation}
The intended result is $w'=h$ and $h'=w$, so Eq.~(\ref{eq:naiveswap}) is off by
a factor $W/H$ on one side and $H/W$ on the other. The aspect ratio of the
result is the intended one multiplied by $(W/H)^2$. Table~\ref{tab:swap} works
through a case with $W/H=2$, where a $400\times100$ box turns into a square
instead of a $100\times400$ box.

\begin{table}[t]
  \centering
  \tabfont
  \caption{Swapping the normalized size instead of the pixel size, for a
  $400\times100$ pixel box in a $1024\times512$ image. Both results have the
  same area as the input.}
  \label{tab:swap}
  \begin{tabular}{@{}lccc@{}}
    \toprule
    & $\bar w$ & $\bar h$ & pixels \\
    \midrule
    input           & $0.3906$ & $0.1953$ & $400\times100$ \\
    correct swap    & $0.0977$ & $0.7813$ & $100\times400$ \\
    normalized swap & $0.1953$ & $0.3906$ & $200\times200$ \\
    \bottomrule
  \end{tabular}
\end{table}

Note that $w'h'=wh$ holds for Eq.~(\ref{eq:naiveswap}) at every aspect ratio, so
the area of the box is preserved and a check on area or on scale will not detect
the mistake. Only the aspect ratio changes. The rule is active mostly on near-square
targets, where the correct swap is close to the identity, so the error replaces
an almost correct target by a clearly wrong one while every aggregate size
statistic remains unchanged. Aerial crops are often square, which hides the
mistake further. We swap on pixel values and renormalize.

\section{Algorithm}

Algorithm~\ref{alg:swap} states the rule as it is applied inside one training
iteration. Lines 4 to 8 run on the three regression targets of Section~2.3, once for the
regression targets of the queries selected by the one-to-many assigner, once
for the supervised DN targets and once for the pseudo-label DN targets. Line 11
is the reliability cost of Eq.~(4). The teacher is unchanged, and so is
inference.

\begin{algorithm}[t]
\tabfont
\caption{Swap-aligned targets in one iteration}
\label{alg:swap}
\begin{algorithmic}[1]
  \REQUIRE student predictions $\{\pp_i\}$, targets $\{\tvec_j\}$, distance $d$
  \STATE run the teacher, threshold its scores, obtain pseudo boxes
  \STATE assign queries to targets with the Hungarian matcher using
         Eq.~(4) as the regression cost
  \STATE fit the two-component GMM to the matched costs, keep the reliable set
  \FOR{each matched pair $(\pp_i,\tvec_j)$ and each DN pair}
    \STATE $\tvec^{\text{sw}} \leftarrow \sigma(\tvec_j)$
           \quad\{exchange on pixels, then renormalize\}
    \STATE $d_0 \leftarrow d(\sg(\pp_i),\tvec_j)$
    \STATE $d_1 \leftarrow d(\sg(\pp_i),\tvec^{\text{sw}})$
    \STATE $\tvec^{*}_j \leftarrow \tvec_j$ if $d_0\le d_1$, else
           $\tvec^{\text{sw}}$
  \ENDFOR
  \STATE accumulate the losses against $\{\tvec^{*}_j\}$ and update the student
  \STATE update the teacher by EMA
\end{algorithmic}
\end{algorithm}

The tie $d_0=d_1$ is resolved in favor of the original encoding, which keeps the
selection inactive when it would not reduce the distance. The comparison returns
an index rather than a differentiable quantity, so the selection creates no
gradient path.

\section{Complete set of ablations}

Table~\ref{tab:all} lists every configuration we trained to completion on the
30\% split. Each row is averaged over three training runs, except the
fixed-reliability-cost control, which was trained once. All runs use
the same detector, the same 180k schedule and the same pseudo-label threshold,
and all of them keep the one-to-many warm-up equal to the total schedule, so
none of the comparisons below is confounded by the Semi-DETR
\cite{zhang2023semidetr} stage switch.

Three groups of runs appear in the table. The rows marked cosine are an earlier
sequence, which covers the full $2\times2$ over the two denoising paths and
allows us to attribute the gain one path at a time. The rows marked wrapped are
the sequence reported in the main paper. The two sequences differ in the angle
penalty
$\varphi$ and cannot be pooled. The final four configurations retain wrapped L1 and align all three regression
targets, then vary only pseudo-label denoising. The configuration with
an early ramp, teacher-score weighting, and encoding-margin weighting is the form
of Oriented Semi-DETR reported in the main paper.

\begin{table*}[t]
  \centering
  \tabfont
  \caption{Complete ablations on DOTA-v2.0 with 30\% labeled images.
  ``Reg.'', ``Sup.\ DN'', and ``Unsup.\ DN'' mark the regression targets to
  which Eq.~(3) is applied. All configurations except the fixed-reliability-cost control use
  the swap-invariant reliability cost of Eq.~(4). The early-ramp, teacher-score, and encoding-margin configuration is
  Oriented Semi-DETR in the main paper.}
  \label{tab:all}
  \setlength{\tabcolsep}{9pt}
  \begin{tabular}{@{}lccccccc@{}}
    \toprule
    Configuration & Reg. & \makecell{Sup.\\DN} & \makecell{Unsup.\\DN}
       & $\varphi$ & $\ap{50}$ & $\ap{75}$ & mAP \\
    \midrule
    \makecell[l]{Fixed reliability\\cost} & & & & cosine & 42.45 & 23.60 & 24.14 \\
    \midrule
    \makecell[l]{Swap-invariant\\cost} & & & & cosine & 42.33 & 23.95 & 24.11 \\
    \makecell[l]{Swap-invariant cost +\\regression target} & \checkmark & & & cosine & 43.65 & 25.54 & 25.50 \\
    \makecell[l]{Swap-invariant cost + regression\\target + supervised DN} & \checkmark & \checkmark & & cosine & 50.34 & 32.33 & 31.20 \\
    \makecell[l]{Swap-invariant cost + regression\\target + pseudo-label DN} & \checkmark & & \checkmark & cosine & 43.25 & 24.90 & 25.07 \\
    \makecell[l]{All three regression\\targets} & \checkmark & \checkmark & \checkmark & cosine & 49.83 & 31.59 & 30.59 \\
    \makecell[l]{All three + angle-aware\\one-to-many alignment} & \checkmark & \checkmark & \checkmark & cosine & 50.44 & 32.08 & 30.97 \\
    \midrule
    \makecell[l]{Swap-invariant cost,\\wrapped L1} & & & & wrapped & 49.15 & 28.68 & 28.69 \\
    \makecell[l]{Wrapped cost +\\supervised DN} & & \checkmark & & wrapped & 50.47 & 30.55 & 30.14 \\
    \makecell[l]{Wrapped cost + supervised DN\\+ regression target} & \checkmark & \checkmark & & wrapped & 51.44 & \textbf{33.54} & 31.89 \\
    \makecell[l]{All three regression targets,\\wrapped L1} & \checkmark & \checkmark & \checkmark & wrapped & \textbf{51.98} & 33.40 & \textbf{31.95} \\
    \makecell[l]{All three regression targets,\\raw L1} & \checkmark & \checkmark & \checkmark & raw & 51.53 & 33.04 & 31.76 \\
    \makecell[l]{Wrapped cost + supervised DN\\+ double-angle head} & & \checkmark & & wrapped & 50.49 & 30.84 & 30.39 \\
    \midrule
    \makecell[l]{All three + early\\pseudo-label DN ramp} & \checkmark & \checkmark & \checkmark & wrapped & 51.84 & 33.57 & 31.96 \\
    \makecell[l]{Early ramp +\\teacher score} & \checkmark & \checkmark & \checkmark & wrapped & 52.08 & \textbf{33.60} & 32.18 \\
    \makecell[l]{Early ramp + teacher score\\+ encoding margin} & \checkmark & \checkmark & \checkmark & wrapped & \textbf{52.11} & 33.51 & \textbf{32.19} \\
    \makecell[l]{Early ramp + teacher score\\+ encoding margin + aspect ratio} & \checkmark & \checkmark & \checkmark & wrapped & 51.51 & 33.02 & 31.73 \\
    \bottomrule
  \end{tabular}
\end{table*}

\subsection{The reliability cost on its own}

The fixed-reliability-cost and swap-invariant-cost controls differ only in the
regression term of the Hungarian cost, with target selection disabled
everywhere. The difference is $-0.12$ $\ap{50}$, $+0.35$ $\ap{75}$, and
$-0.03$ mAP. The swap-invariant cost gains at the strict threshold but loses marginally
at the loose threshold, so this control does not establish a final-accuracy
gain by itself. The comparison is unbalanced because the fixed-cost control was
run once and the swap-invariant-cost control three times. Both operate near $42$
  $\ap{50}$, far below the $50.63$ reached by Semi-DETR
  \cite{zhang2023semidetr}, because the metric
control replaces plain L1 by the $\pi$-periodic cosine form everywhere while
leaving every target unchanged. That substitution alone costs about eight
points, so the pair probes a weak operating point.

This is the reason the main paper presents Eq.~(4) as mechanism evidence,
supported by the diagnostic of Section~\ref{sec:diag}, and keeps it enabled in
every row of Table~2 rather than claiming an AP gain for it.

\subsection{Which denoising path matters}

The cosine configurations with a swap-invariant cost and regression target form a
$2\times2$ study over the two DN paths.
Adding supervised DN gives $+6.69$ $\ap{50}$ and $+6.79$ $\ap{75}$ over the
regression-target-only configuration. Pseudo-label DN alone changes these
metrics by $-0.40$ and $-0.64$ points. Adding it on top of supervised DN changes
them by $-0.51$ and $-0.74$ points, respectively.

The wrapped-L1 configurations lead to the same qualitative conclusion. Adding
supervised DN to the swap-invariant cost raises $\ap{75}$ by $1.87$ points, and adding the
regression target raises it by another $2.99$ points. Adding pseudo-label DN
target selection changes $\ap{75}$ by $-0.14$ points. Because the two angle
penalties were trained in different batches, these values are not pooled. The
pseudo-label DN path remains enabled because the wrapped-L1 configuration with
all three regression targets aligned improves $\ap{50}$ by $0.54$ points and mAP by $0.06$ points over
the configuration without that path. It is active only during the last 20k
iterations of the 180k schedule. The early-ramp study below lengthens this
window.

\subsection{Choice of the angle penalty}
\label{sec:phi}

The cosine, wrapped-L1 and raw-L1 configurations that align all three
regression targets differ only in
$\varphi$. Relative to the cosine form, wrapped L1
gains $2.15$ $\ap{50}$, $1.81$ $\ap{75}$, and $1.36$ mAP; raw L1 gains $1.70$,
$1.45$, and $1.17$ points. The ordering is the same for all three metrics, so
we use wrapped L1.

The caveat of Section~A applies in full. Since $\lambda$ stays at $1$, the three
rows differ in the effective weight of the angle term as well as in the shape of
the penalty, and the cosine form assigns four times as much as the other two at
a $45^\circ$ disagreement. The comparison identifies the better configuration;
it does not isolate the function shape. A clean shape study would rescale
$\lambda$ so that the three penalties agree at a chosen reference angle, and we
have not run it.

The cosine configuration with angle-aware one-to-many alignment adds $0.61$
$\ap{50}$, $0.49$ $\ap{75}$, and $0.38$ mAP over the corresponding cosine
configuration. We did not rerun this addition under wrapped L1, and it is not
part of Oriented Semi-DETR.

\subsection{A representation-level alternative}

The double-angle-head configuration replaces the scalar angle output by the
continuous vector $(\cos2\theta,\sin2\theta)$, decoded back to the same
five-parameter interface. Apart from that change, it matches the wrapped-L1
configuration with supervised DN. It gains $+0.02$ $\ap{50}$, $+0.29$
$\ap{75}$, and $+0.25$ mAP.

The outcome is what the analysis of Section~B predicts. The double-angle vector
removes the discontinuity of the scalar angle at the wrap point, since $\theta$
and $\theta+\pi$ map to the same vector. It does not remove the swap ambiguity,
because $\sigma$ negates the vector and exchanges $w$ and $h$ at the same time.
The ambiguity is a joint property of $(w,h,\theta)$, so no re-encoding of the
angle alone can be invariant under it. This separates our rule from the angle
encoding literature, CSL \cite{yang2020csl} and PSC \cite{yu2023psc} among
others, and it is also why the Gaussian losses of Section~2.6 are the relevant
point of comparison: a Gaussian is a function of the rectangle itself.
The head was not recalibrated after this change, so this is not a conclusive
comparison of output representations.

\subsection{The pseudo-label denoising schedule}
\label{sec:sched}

The final four configurations retain the aligned wrapped-L1 design and change
only how pseudo-label denoising is opened and weighted. The first moves the
start from 160k to 120k and adds a 10k linear ramp, with no per-query weight.
The second multiplies the regression weight of each denoising query by the
teacher score $q$ of its pseudo box. The third additionally multiplies the
angle term by
\begin{equation}
  m = \frac{|d(\pp,\bb)-d(\pp,\sigma(\bb))|}
           {d(\pp,\bb)+d(\pp,\sigma(\bb))+\varepsilon},
  \label{eq:margin}
\end{equation}
with $\varepsilon=10^{-6}$, which is large when the two encodings are clearly
distinguishable and small when the prediction lies between them. The fourth
configuration redistributes angle weight by the target aspect ratio $r=\max(w,h)/\min(w,h)$, using the
mass-conserving multiplier $u=1+(r^2-1)/(r^2+1)$ normalized to keep the total
angle weight fixed.

The encoding-margin and aspect-ratio controls do not establish a uniform gain over
teacher-score weighting. Both reduce the angle emphasis for near-square targets.
For aspect-ratio weighting, this follows directly because $u$ increases with
$r$. The diagnostic in Section~\ref{sec:diag} shows the opposite distribution
for the ambiguity itself: at 180k it occurs for $22.1\%$ of near-square targets and
$1.2\%$ of elongated targets. Aspect-ratio weighting therefore transfers angle
supervision away from the group where the ambiguity is most common. Because the
redistribution conserves total mass, this transfer is not offset elsewhere.

Averaged over pseudo-label DN positives, the encoding margin is $0.855$, so it
reduces angle weight by about six percent relative to spatial weight. The
aspect-ratio multiplier is bounded in $[1,2)$ before normalization and produces
about a $1.5\times$ tilt between near-square and elongated targets on the
observed distribution. These controls are retained as component analyses rather
than presented as separate methods.

\section{Results across label ratios}
\label{sec:label-ratios}

The main paper reports the same DOTA-v2.0 \cite{xia2018dota,ding2022dotav2} partial-label protocol of
Semi-DETR \cite{zhang2023semidetr} at
10\%, 20\%, and 30\% labeled images. Table~\ref{tab:ratios} gives the mean
test-dev metrics for DINO \cite{zhang2023dino}, Semi-DETR \cite{zhang2023semidetr},
Sparse Semi-DETR \cite{shehzadi2024sparse}, and Oriented Semi-DETR.
The three component settings below the main methods are controlled parts of
Oriented Semi-DETR, and we do not name them as separate methods. They successively add an
early pseudo-label DN ramp, teacher-score weighting, and encoding-margin weighting.

\begin{table*}[t]
  \centering
  \tabfont
  \caption{Mean DOTA-v2.0 test-dev performance across label ratios and in the
  fully labeled setting. Oriented Semi-DETR uses the early ramp, teacher score,
  and encoding margin.}
  \label{tab:ratios}
  \setlength{\tabcolsep}{2.6pt}
  \begin{tabular}{@{}lcccccccccccc@{}}
    \toprule
    \multirow{2}{*}{Method or component setting}
      & \multicolumn{3}{c}{10\% labeled}
      & \multicolumn{3}{c}{20\% labeled}
      & \multicolumn{3}{c}{30\% labeled}
      & \multicolumn{3}{c}{Fully labeled} \\
    \cmidrule(lr){2-4}\cmidrule(lr){5-7}\cmidrule(lr){8-10}\cmidrule(lr){11-13}
      & $\ap{50}$ & $\ap{75}$ & mAP
      & $\ap{50}$ & $\ap{75}$ & mAP
      & $\ap{50}$ & $\ap{75}$ & mAP
      & $\ap{50}$ & $\ap{75}$ & mAP \\
    \midrule
    DINO & 40.72 & 21.76 & 22.49 & 46.31 & 24.94 & 25.82 & 48.11 & 26.71 & 27.16
      & 49.11 & 26.91 & 27.51 \\
    Semi-DETR & 43.94 & 25.62 & 25.82 & 48.01 & 28.10 & 28.10 & 50.63 & 29.47 & 29.60
      & 51.11 & 31.50 & 30.81 \\
    Sparse Semi-DETR & 44.16 & 25.49 & 25.83 & 48.38 & 28.28 & 28.51 & 50.24 & 29.40 & 29.53
      & 51.71 & 32.26 & 31.22 \\
    \midrule
    Oriented Semi-DETR & \textbf{45.92} & \textbf{27.77} & \textbf{27.61}
      & \textbf{49.89} & \textbf{31.67} & \textbf{30.69}
      & \textbf{52.11} & \textbf{33.51} & \textbf{32.19}
      & \textbf{53.16} & \textbf{35.07} & \textbf{33.23} \\
    \midrule
    \makecell[l]{Early pseudo-label DN\\ramp} & 45.56 & \textbf{27.88} & 27.52
      & 49.19 & \textbf{32.15} & \textbf{30.69}
      & 51.84 & 33.57 & 31.96
      & 52.79 & 34.86 & 33.15 \\
    \makecell[l]{Early ramp +\\teacher score} & \textbf{45.92} & 27.77 & \textbf{27.65}
      & 49.64 & 31.95 & 30.50
      & 52.08 & \textbf{33.60} & 32.18
      & 53.02 & 34.50 & 32.99 \\
    \makecell[l]{Early ramp + teacher\\score + encoding margin} & \textbf{45.92} & 27.77 & 27.61
      & \textbf{49.89} & 31.67 & \textbf{30.69}
      & \textbf{52.11} & 33.51 & \textbf{32.19}
      & \textbf{53.16} & \textbf{35.07} & \textbf{33.23} \\
    \bottomrule
  \end{tabular}
\end{table*}

At 10\%, teacher-score weighting gives the highest mAP ($27.65$), the early
ramp gives the highest $\ap{75}$ ($27.88$), and teacher-score and encoding-margin
weighting tie in $\ap{50}$ ($45.92$). At 20\%, encoding-margin weighting has the
highest $\ap{50}$ ($49.89$), the early ramp has the highest $\ap{75}$ ($32.15$),
and the early-ramp and encoding-margin configurations tie in mAP ($30.69$). At
30\%, the corresponding values are already reported in the main ablation. In
the fully labeled setting the ordering is cleaner: the complete configuration
has the highest $\ap{50}$ ($53.16$), $\ap{75}$ ($35.07$) and mAP ($33.23$), and
the encoding margin recovers the $\ap{75}$ that teacher-score weighting alone loses
($34.50$ against $34.86$ for the ramp alone). Across the partial splits,
however, no component setting dominates every metric at every label ratio. We
report the nested component study explicitly and use Oriented Semi-DETR only for
the complete method.

Moving from the 30\% split to the fully labeled setting yields about one point
of mAP, less than the step from 20\% to 30\%. Two properties of the protocol
limit it. The 180k schedule is inherited unchanged from the partial splits and
is short for a labeled set three times the size, so the extra annotation is
seen fewer times per image. The unlabeled pool also changes character: the
test-challenge images carry more noise and more hard cases than the trainval set, even
after we drop the frames under heavy cloud or fog and the frames whose
foreground is almost black. A longer schedule and a cleaner unlabeled pool are
the two direct ways to test whether the gap is a property of the method or of
the budget, and we have run neither.
\begin{table*}[t]
  \centering
  \tabfont
  \caption{Class-wise $\ap{50}$ means for the three Oriented Semi-DETR component
  settings at 10\% and 20\% labeled images and in the fully labeled setting. The
  official output provides only class-wise $\ap{50}$.}
  \label{tab:class-ratios}
  \setlength{\tabcolsep}{2.5pt}
  \begin{tabular}{@{}lrrrrrrrrr@{}}
    \toprule
    \multirow{2}{*}{Class} & \multicolumn{3}{c}{10\% labeled}
      & \multicolumn{3}{c}{20\% labeled} & \multicolumn{3}{c}{Fully labeled} \\
    \cmidrule(lr){2-4}\cmidrule(lr){5-7}\cmidrule(lr){8-10}
      & \makecell{Early\\ramp} & \makecell{+ Teacher\\score} & \makecell{+ Encoding\\margin}
      & \makecell{Early\\ramp} & \makecell{+ Teacher\\score} & \makecell{+ Encoding\\margin}
      & \makecell{Early\\ramp} & \makecell{+ Teacher\\score} & \makecell{+ Encoding\\margin} \\
    \midrule
    plane & 77.01 & 77.18 & 76.92 & 77.19 & 78.65 & 78.88 & 79.41 & 79.16 & 78.98 \\
    baseball-diamond & 54.56 & 54.61 & 53.78 & 53.55 & 52.26 & 53.85 & 49.62 & 50.53 & 52.92 \\
    bridge & 45.37 & 45.04 & 44.36 & 48.67 & 48.77 & 48.62 & 48.77 & 48.63 & 49.01 \\
    ground-track-field & 61.92 & 62.42 & 62.74 & 65.82 & 64.66 & 65.43 & 67.17 & 67.90 & 67.28 \\
    small-vehicle & 36.24 & 35.19 & 36.35 & 39.30 & 39.67 & 38.00 & 41.41 & 40.83 & 40.88 \\
    large-vehicle & 41.22 & 41.73 & 40.99 & 42.44 & 43.80 & 43.71 & 45.23 & 44.97 & 45.72 \\
    ship & 44.13 & 43.54 & 43.50 & 44.54 & 47.93 & 46.78 & 50.94 & 50.97 & 51.01 \\
    tennis-court & 76.40 & 76.34 & 77.20 & 74.49 & 74.18 & 73.66 & 82.65 & 82.45 & 82.26 \\
    basketball-court & 54.75 & 53.33 & 54.85 & 58.13 & 57.27 & 56.11 & 63.18 & 62.48 & 62.60 \\
    storage-tank & 59.24 & 59.14 & 59.12 & 59.72 & 59.62 & 59.50 & 63.53 & 62.38 & 62.28 \\
    soccer-ball-field & 32.48 & 34.50 & 32.84 & 39.63 & 41.00 & 40.37 & 43.43 & 45.97 & 45.61 \\
    roundabout & 50.40 & 50.85 & 49.93 & 53.08 & 52.52 & 52.62 & 57.52 & 57.10 & 57.98 \\
    harbor & 39.64 & 39.45 & 39.73 & 38.32 & 38.60 & 38.36 & 43.96 & 43.90 & 44.16 \\
    swimming-pool & 50.31 & 51.10 & 50.50 & 51.80 & 52.64 & 52.39 & 53.84 & 53.99 & 53.98 \\
    helicopter & 37.23 & 38.82 & 42.19 & 41.62 & 47.07 & 51.13 & 58.32 & 60.02 & 58.85 \\
    container-crane & 4.37 & 7.59 & 7.32 & 12.62 & 11.66 & 14.02 & 7.29 & 7.75 & 8.80 \\
    airport & 47.87 & 50.21 & 50.28 & 78.06 & 76.71 & 76.71 & 85.68 & 84.49 & 85.67 \\
    helipad & 6.87 & 5.60 & 4.01 & 6.42 & 6.49 & 7.82 & 8.30 & 10.89 & 8.86 \\
    \bottomrule
  \end{tabular}
\end{table*}

\section{The diagnostic in full}
\label{sec:diag}

\subsection{Protocol}

We load a training checkpoint, run the teacher on 128 unlabeled images and
keep the pseudo boxes above the usual score threshold. The raw student queries
of the same forward pass are then matched to those pseudo boxes twice. The first
pass uses the fixed regression cost, the second uses Eq.~(4). The
classification and rotated IoU terms are shared between the two passes, and each
pass runs its own Hungarian assignment and fits its own two-component GMM to the
matched costs of that batch, exactly as training does. A record is one matched
pseudo box, so every percentage below is taken over matched pairs. We report two checkpoints,
20k and 180k, to show the effect at the start and at the end of training.

\subsection{Overall behavior}

Table~\ref{tab:diagfull} gives the complete counts. ``Sigma'' is the fraction of
matched pairs for which the swapped encoding of the pseudo box is closer to the
matched query than the teacher's own encoding. It is reported for both matchers, and the
gap between the two columns is the first result: the fixed matcher selects a
swapped query on its own for $0.09\%$ of the pairs at 20k, against $16.89\%$ for
the swap-invariant matcher. Beyond mispricing the ambiguous pairs, the fixed
cost diverts the assignment away from them, so the filter never evaluates them. This is
also why the diagnostic must rerun the assignment rather than only rescoring a
fixed matching.

\begin{table*}[t]
  \centering
  \tabfont
  \caption{Fixed versus swap-invariant reliability cost on teacher outputs from 128
  unlabeled images. ``Sigma'' is the fraction of matched pairs whose swapped encoding
  is closer, under each matcher. ``Assign.\ flip'' counts pairs whose matched
  query changes, ``Filter flip'' counts pairs whose keep or drop decision
  changes. ``Keep'' is the fraction of pseudo boxes the GMM retains. The last
  two columns split the filter flips into boxes kept only under the swap-invariant cost
  and boxes kept only under the fixed cost.}
  \label{tab:diagfull}
  \begin{tabular}{@{}lrcccccccc@{}}
    \toprule
    \multirow{2}{*}{Step} & \multirow{2}{*}{Pairs}
      & \multicolumn{2}{c}{Sigma} & \multirow{2}{*}{\makecell{Assign.\\flip}}
      & \multirow{2}{*}{\makecell{Filter\\flip}}
      & \multicolumn{2}{c}{Keep} & \multirow{2}{*}{\makecell{Swap\\only}}
      & \multirow{2}{*}{\makecell{Fixed\\only}} \\
    \cmidrule(lr){3-4}\cmidrule(lr){7-8}
      & & fixed & swap & & & fixed & swap & & \\
    \midrule
    20k  & 4671 & 0.09\% & 16.89\% & 18.88\% & 7.02\% & 35.54\% & 36.74\% & 4.11\% & 2.91\% \\
    180k & 3826 & 0.05\% & 3.89\%  & 4.89\%  & 2.88\% & 36.23\% & 36.02\% & 1.33\% & 1.54\% \\
    \bottomrule
  \end{tabular}
\end{table*}

The second result is the contrast between the flip columns and the keep columns.
At 20k the net keep rate rises by $1.20$ points while $7.02\%$ of the individual
decisions change, split almost evenly into $4.11\%$ gained and $2.91\%$ lost. At
180k the net keep rate falls by $0.21$ points and $2.88\%$ of the decisions still
change. The swap-invariant cost retains a comparable number of pseudo boxes and a different
selection of them. Any argument for it therefore must concern which boxes it
keeps.

Restricting attention to the pairs that the swap-invariant matcher assigns to a
swapped query makes the effect more explicit. There are 789 such pairs at 20k,
of which the fixed pipeline keeps $1.39\%$ and the swap-invariant pipeline
keeps $16.73\%$. At 180k there
are 149, kept at $1.34\%$ and $12.75\%$. These are the pairs on which a
perfectly localized student prediction is priced as a failure, and the fixed
pipeline discards almost all of them.

Both rates fall by roughly a factor of four between 20k and 180k. The teacher
becomes more consistent as it converges, so the ambiguity is a property of the
early pseudo-label distribution more than of the converged one. A measurement
taken on a finished model therefore understates it, since it is largest while
the pseudo-label distribution is being formed and has mostly disappeared by the
time training ends.

\subsection{Aspect ratio}

Table~\ref{tab:arfull} splits the same pairs by the aspect ratio of the pseudo
box at both checkpoints. The effect is confined to near-square boxes at both. In
the aspect-ratio range $(1,1.3]$ at 20k, the swapped encoding is closer for $43.92\%$ of the
pairs, $45.49\%$ of the assignments change, and the keep rate rises from
$28.63\%$ to $44.90\%$, a gain of $16.27$ points. In the range $(2,\infty)$ the
first two numbers fall to $7.05\%$ and $8.62\%$ and the keep rate moves by
$1.49$ points in the other direction. At 180k the near-square range still shows
$22.09\%$ sigma and $11.64\%$ filter flips, but the keep rates have converged, so
what remains is a substitution of one set of pseudo boxes for another.

\begin{table}[t]
  \centering
  \tabfont
  \caption{The diagnostic split by aspect ratio at both checkpoints. ``Keep''
  is given under the fixed and the swap-invariant cost.}
  \label{tab:arfull}
  \setlength{\tabcolsep}{3.2pt}
  \begin{tabular}{@{}llrcccc@{}}
    \toprule
    Step & \makecell[l]{Aspect\\ratio} & $n$ & Sigma
      & \makecell{Assign.\\flip} & \makecell{Filter\\flip}
      & \makecell{Keep\\fixed / swap} \\
    \midrule
    \multirow{3}{*}{20k}
      & $(1,1.3]$    &  510 & 43.92\% & 45.49\% & 18.63\% & 28.63\% / 44.90\% \\
      & $(1.3,2]$    & 1283 & 28.22\% & 31.33\% &  8.26\% & 27.36\% / 28.60\% \\
      & $(2,\infty)$ & 2878 &  7.05\% &  8.62\% &  4.41\% & 40.41\% / 38.92\% \\
    \midrule
    \multirow{3}{*}{180k}
      & $(1,1.3]$    &  335 & 22.09\% & 25.97\% & 11.64\% & 41.49\% / 41.19\% \\
      & $(1.3,2]$    & 1109 &  4.24\% &  5.50\% &  3.16\% & 30.21\% / 29.58\% \\
      & $(2,\infty)$ & 2382 &  1.18\% &  1.64\% &  1.51\% & 38.29\% / 38.29\% \\
    \bottomrule
  \end{tabular}
\end{table}

This is the expected shape. For an elongated box the two encodings differ
strongly in $w$ and $h$, so the size terms of Eq.~(\ref{eq:supdist}) separate
them and the teacher's encoding is almost always the closer one. For a near-square
box the size terms nearly cancel and the decision rests on the angle alone.
Near-square objects are common in aerial imagery: storage tanks, roundabouts,
swimming pools and vehicles seen from above, all represented among the DOTA
categories \cite{xia2018dota,ding2022dotav2}. A mechanism that would be marginal
in general object detection is important here.

\subsection{Class composition}

Table~\ref{tab:cls} reports the five classes that contribute most pairs. The
clearest case is the swimming pool, which is close to square in plan view: at
20k the swapped encoding is closer for $57.07\%$ of its pairs and the keep rate
rises from $21.20\%$ to $51.63\%$. Elongated classes behave as
Table~\ref{tab:arfull} predicts. Large vehicles show $7.06\%$ sigma at 20k and
$0.79\%$ at 180k, and their keep rate changes only slightly.

\begin{table}[t]
  \centering
  \tabfont
  \caption{The diagnostic for the five classes contributing the most matched
  pairs.}
  \label{tab:cls}
  \setlength{\tabcolsep}{2.4pt}
  \begin{tabular}{@{}llrcccc@{}}
    \toprule
    Step & Class & $n$ & Sigma & \makecell{Assign.\\flip}
      & \makecell{Filter\\flip} & \makecell{Keep\\fixed / swap} \\
    \midrule
    \multirow{5}{*}{20k}
      & ship          &   32 & 34.38\% & 37.50\% &  9.38\% & 37.50\% / 46.88\% \\
      & small vehicle &  200 & 16.00\% & 16.50\% &  5.00\% & 55.00\% / 56.00\% \\
      & large vehicle &  680 &  7.06\% &  8.97\% &  1.62\% & 44.41\% / 43.38\% \\
      & swimming pool &  184 & 57.07\% & 60.33\% & 33.70\% & 21.20\% / 51.63\% \\
      & storage tank  & 2173 & 13.99\% & 16.43\% &  6.26\% & 36.45\% / 34.70\% \\
    \midrule
    \multirow{5}{*}{180k}
      & ship          &   31 &  3.23\% &  3.23\% &  3.23\% & 48.39\% / 51.61\% \\
      & small vehicle &   94 &  2.13\% &  2.13\% &  4.26\% & 53.19\% / 53.19\% \\
      & large vehicle &  508 &  0.79\% &  0.98\% &  0.59\% & 43.70\% / 43.50\% \\
      & swimming pool &  225 & 26.22\% & 31.11\% & 15.56\% & 38.22\% / 36.89\% \\
      & storage tank  & 2006 &  2.09\% &  3.24\% &  2.39\% & 32.75\% / 32.45\% \\
    \bottomrule
  \end{tabular}
\end{table}

\subsection{What the diagnostic does not show}

Four limits are worth stating explicitly.

The sample is 128 unlabeled images. Pairs from the same image are strongly
correlated, so the counts of $4671$ and $3826$ cannot be treated as independent
observations and we do not attach confidence intervals to any of these rates.

The class composition is uneven. Storage tanks alone are $46.5\%$ of the pairs
at 20k and $52.4\%$ at 180k, and $59.7\%$ of the near-square pairs at 180k are
swimming pools. Aspect ratio and class are therefore confounded, and
Tables~\ref{tab:arfull} and \ref{tab:cls} should be read as two views of the
same population rather than as independent evidence.

The evaluation server returns per-class results at $\ap{50}$ only. We cannot
connect the classes of Table~\ref{tab:cls} to per-class $\ap{75}$, so the
diagnostic cannot be checked class by class against the accuracy gain.

Finally, the diagnostic measures routing and filtering at two fixed checkpoints.
It identifies the error the fixed cost makes and how often it occurs; it does
not by itself establish a gain in final accuracy. The accuracy claims rest on
the training runs of Table~\ref{tab:all}.

\section{Hyperparameters}

Table~\ref{tab:hyper} lists the settings shared by the runs in
Table~\ref{tab:all}. They use the Semi-DETR-style training pipeline
\cite{zhang2023semidetr}; the shared entries were not tuned for this work. The only
configuration-specific exception is the
pseudo-label DN schedule, which is reported separately below because the
legacy controls and the complete Oriented Semi-DETR configuration use different
start times.

\begin{table}[t]
  \centering
  \tabfont
  \caption{Training settings. All entries are shared unless a
  configuration-specific pseudo-label DN schedule is stated explicitly.}
  \label{tab:hyper}
  \begin{tabular}{@{}ll@{}}
    \toprule
    Backbone & ResNet-50 \cite{he2016resnet}, ImageNet initialized \\
    Detector & DINO \cite{zhang2023dino}, 900 queries, 100 DN queries \\
    Optimizer & AdamW, lr $10^{-4}$, weight decay $10^{-4}$ \\
    Backbone lr & $0.1\times$ the base rate \\
    Schedule & 180k iterations, lr $\div10$ at 120k and 160k \\
    Batch & 1 labeled and 1 unlabeled image per GPU \\
    Train scale & long side 1333, short side sampled in $[1024,1200]$ \\
    Test scale & $1024\times1024$ \\
    Augmentation & random flip and rotation, weak for the teacher \\
    Teacher & EMA, momentum $0.999$ \\
    Pseudo labels & score threshold $0.4$ \\
    Unsupervised weight & $0.5$ \\
    O2M warm-up & 180k, equal to the schedule \\
    \makecell[l]{Pseudo-label DN\\(legacy)} & 160k start \\
    \makecell[l]{Pseudo-label DN\\(complete method)} & 120k start + 10k ramp \\
    Matching cost & focal $2$, regression $5$, rotated IoU $2$ \\
    Losses & box $5$, rotated IoU $2$, $\lambda=1$ \\
    Seed & 42 \\
    \bottomrule
  \end{tabular}
\end{table}

The warm-up length deserves a comment because it decides which Semi-DETR
\cite{zhang2023semidetr} stage the runs are in. Semi-DETR uses a one-to-many
assignment first and switches to one-to-one after the warm-up. In our protocol,
we set the warm-up to 180k, equal to the total schedule, so every run stays in
the first stage and the one-to-one branch, together with the cross-view query
consistency associated with that stage, is not activated. This is our controlled
experimental choice, rather than a claim about a universal Semi-DETR default;
it keeps the comparison in Table~\ref{tab:all} on the same stage. Consequently,
the results here speak to the one-to-many stage, while the effect of the proposed
rule in the one-to-one stage remains untested.

\section{Protocol and open points}

All experiments use DOTA-v2.0 \cite{xia2018dota,ding2022dotav2} and are
evaluated by the official server on test-dev. The 10\%, 20\% and 30\% splits
follow the partial annotation protocol of Semi-DETR \cite{zhang2023semidetr},
fold 1, sampled from a trainval set that merges the official train and val
sets. The fully labeled setting keeps the entire trainval set as labeled data and uses
the test-challenge images as unlabeled data, after we drop the frames under
heavy cloud or fog and the frames whose foreground is almost black. It starts
the unlabeled loss at 20k instead of 10k, and every other hyperparameter,
including the 180k schedule, the one-to-many warm-up length and the
pseudo-label threshold, is the one used for the partial splits.

Every run uses seed 42, so the runs behind each mean differ only through
nondeterminism that the seed does not control, mainly nondeterministic CUDA
kernels and data loading order. The means we report therefore average out
the nondeterminism that remains at a fixed seed. They do not support a significance
test. A multi-seed study is the appropriate way to strengthen the statistical
claims, and we have not run one.

Two points remain open. The isolation of the reliability cost rests on an
unbalanced comparison between the fixed-reliability-cost control and the
swap-invariant-cost control, and would be worth rerunning on equal footing.
The angle penalty comparison of
Section~\ref{sec:phi} confounds the shape of $\varphi$ with its scale, and a
rescaled control would separate them.

\bibliographystyle{IEEEbib}
\bibliography{refs}